\documentclass[letterpaper,12pt]{article}
\usepackage{tabularx} 
\usepackage{amsmath}  
\usepackage{graphicx} 
\usepackage[space]{grffile}
\usepackage[margin=1in,letterpaper]{geometry} 
\usepackage{cite} 
\usepackage{float}
\PassOptionsToPackage{hyphens}{url}\usepackage[final]{hyperref}
\hypersetup{
	colorlinks=true,       
	linkcolor=blue,        
	citecolor=blue,        
	filecolor=magenta,     
	urlcolor=blue         
}
\usepackage{blindtext}
\usepackage{indentfirst}
\usepackage{bm}
\usepackage[table]{xcolor}

\usepackage{fancyhdr}
\usepackage{extramarks}

\usepackage{listings}
\usepackage{color}
\usepackage{caption}
\usepackage{xcolor}
\usepackage[most]{tcolorbox}
\usepackage{inconsolata}

\definecolor{dkgreen}{rgb}{0,0.6,0}
\definecolor{gray}{rgb}{0.97,0.97,0.97}
\definecolor{darkgray}{rgb}{0.3,0.3,0.3}
\definecolor{mauve}{rgb}{0.58,0,0.82}

\newtcblisting[auto counter]{out}{sharp corners, 
    fonttitle=\bfseries, colframe=darkgray, listing only, 
    listing options={basicstyle=\ttfamily}, 
    title=Output}

\newtcblisting[auto counter]{outal1}{ colframe=darkgray, listing only, 
    listing options={basicstyle=\ttfamily}}

\lstdefinestyle{pycode}{
	frame=single,
	backgroundcolor=\color{gray},
  	language=Python,
  	aboveskip=3mm,
  	belowskip=3mm,
  	numbers=left,
  	showstringspaces=false,
  	columns=flexible,
  	basicstyle={\small\ttfamily},
  	numberstyle=\tiny\color{black},
  	keywordstyle=\color{blue},
  	commentstyle=\color{dkgreen},
  	stringstyle=\color{mauve},
  	breaklines=true,
  	breakatwhitespace=true,
  	tabsize=3
}

\lstdefinestyle{bash_code}{
	frame=single,
	backgroundcolor=\color{gray},
  	language=bash,
  	aboveskip=3mm,
  	belowskip=3mm,
  	numbers=left,
  	showstringspaces=false,
  	columns=flexible,
  	basicstyle={\small\ttfamily},
  	numberstyle=\tiny\color{black},
  	keywordstyle=\color{blue},
  	commentstyle=\color{dkgreen},
  	stringstyle=\color{mauve},
  	breaklines=true,
  	breakatwhitespace=true,
  	tabsize=3
}

\usepackage{tikz}
\usetikzlibrary{shapes.geometric, arrows}
\usetikzlibrary{positioning}
\usetikzlibrary{arrows.meta}

\tikzstyle{mainblock} = [rectangle, minimum width = 3cm, minimum height = 2cm, text centered, draw = black, fill = white]
\tikzstyle{block} = [rectangle, minimum width = 3cm, minimum height = 3cm, text centered, draw = black, fill = white]
\tikzstyle{arrow} = [->, >=stealth]

\usepackage{titlesec}
\titleformat{\section}[block]{\Large\bfseries\filcenter}{}{1em}{}

\begin{document}

\begin{titlepage}

\begin{center}
\line(1,0){450}\\[1mm]
\huge{\textbf{Self-Supervised Deep Learning \\\ for Robotic Grasping}}\\
\line(1,0){450}\\[15mm]
\Large{Final Year Project Report}\\
\Large{by}\\[5mm]
\Large{\textbf{Danyal Saqib}}\\[30mm]
\Large{In Partial Fulfillment}\\
\Large{Of the Requirements for the degree}\\
\Large{Bachelors of Engineering in Electrical Engineering (BEE)}\\[20mm]
\vfill
\Large{School of Electrical Engineering and Computer Science}\\[5mm]
\Large{National University of Sciences and Technology}\\[5mm]
\Large{Islamabad, Pakistan}\\[5mm]
\Large{2021}
\end{center}
\pagebreak
\end{titlepage}

\cleardoublepage
\pagebreak

\renewcommand{\sectionmark}[1]{\markright{#1}}
\fancyhf{}
\pagenumbering{roman}
\cfoot{\fancyplain{}{\thepage}}
\section*{DECLARATION}
\vspace{6mm}
{\large{ \noindent We hereby declare that this project report entitled ``Self-Supervised Deep Learning for Robotic Grasping" submitted to the ``Department of Electrical Engineering", is a record of an original work done by us under the guidance of Supervisor ``Dr. Wajahat Hussain" and that no part has been plagiarized without citations. Also, this project work is submitted in the partial fulfillment of the requirements for the degree of Bachelor of Engineering in Electrical Engineering.\\\\\\
\begin{tabular}{p{0.4\textwidth} p{0.4\textwidth}}
\textbf{Team Members} & \textbf{Signature}\\\\
Danyal Saqib & \raisebox{-4mm}{\includegraphics[scale=0.1]{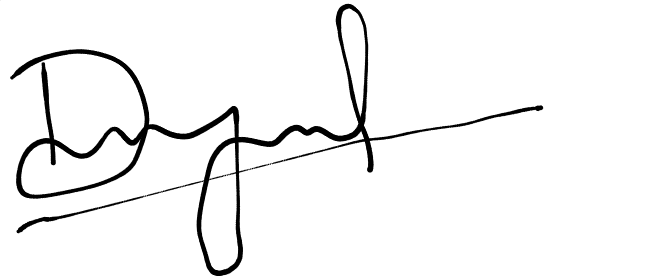}}\\[20mm]
\textbf{Supervisor} & \textbf{Signature}\\\\
Dr. Wajahat Hussain\\[20mm]
\end{tabular}\\
\begin{tabular}{p{0.6\textwidth} p{0.4\textwidth}}
\textbf{Date:}\\\\
28/5/2021\\\\\\
\textbf{Place:}\\\\
SEECS, NUST, Islamabad, Pakistan
\end{tabular}}}
\pagebreak

\fancyhf{}
\cfoot{\fancyplain{}{\thepage}}
\vspace*{60mm}
\section*{DEDICATION}
\vspace{6mm}
\begin{center}
{\large{\noindent Dedicated to my parents for being with me every step of the way, and for believing in me when even I did not.}}
\end{center}
\pagebreak

\fancyhf{}
\cfoot{\fancyplain{}{\thepage}}
\vspace*{40mm}
\section*{ACKNOWLEDGEMENTS}
\vspace{6mm}
{\large{\noindent Foremost, I would like to thank my supervisor Dr. Wajahat Hussain. He gave me countless opportunities to work with him on exciting projects, from giving me the chance to be a Teaching Assistant, to letting us work on this novel research project. In addition, he was extremely helpful throughout the process, and I could not have hoped for a better advisor for this project.\\\\
I would also like to express my gratitude for my team members for this project. While the project had to be split into two distinct modules due to the global pandemic, for the whole research to come together, we still had to co-operate throughout. I am thankful to all of my team members for making this a smooth process. My special thanks to Mr. Salman Nasir for sitting with me through countless sleepless nights. The integration of our two modules would not have been possible without you.\\\\
Finally, I am of course indebted to my family for their constant support through not only this project, but throughout the duration of my degree. I would never be where I am today without them.}}
\pagebreak

\fancyhf{}
\cfoot{\fancyplain{}{\thepage}}
\begin{center}
\Large{Final Year Project Report}
\end{center}
\vspace*{1cm}
\tableofcontents
\cleardoublepage
\pagebreak

\fancyhf{}
\cfoot{\fancyplain{}{\thepage}}
\begin{center}
\Large{Final Year Project Report}
\end{center}
\vspace*{1cm}
\listoffigures
\vspace*{1cm}
\listoftables
\pagebreak

\pagestyle{fancy}
\renewcommand{\sectionmark}[1]{\markright{#1}}
\lhead{\fancyplain{}{Self-Supervised Deep Learning}}
\rhead{\fancyplain{}{\rightmark }} 
\cfoot{\fancyplain{}{\thepage}}
\cleardoublepage
\pagenumbering{arabic}
\vspace*{40mm}
\section{Chapter 1: Abstract}
Learning Based Robot Grasping currently involves the use of labeled data. This approach has two major disadvantages. Firstly, labeling data for grasp points and angles is a strenuous process, so the dataset remains limited. Secondly, human labeling is prone to bias due to semantics.\\

In order to solve these problems we propose a simpler self-supervised robotic setup, that will train a Convolutional Neural Network (CNN). The robot will label and collect the data during the training process. The idea is to make a robot that is less costly, small and easily maintainable in a lab setup. The robot will be trained on a large data set for several hundred hours and then the trained Neural Network can be mapped onto a larger grasping robot.
\pagebreak

\section{Chapter 2: Introduction}
The basic problem our project is trying to solve is dynamic robot grasping. Currently, solving this particular problem is a critical challenge in robotics. Robots are already very good at working in tightly monitored, closed, and predictable environments such as assembly lines in factories. This is because in such environments, the robotic hand has to pick up the same object, with the same geometric shape, from the same position over and over again. As such, the problem is very much static in that the robotic hand can be built and programmed for that specific task.\\

The idea to use robot hand-eye coordination and visual servoing dates back to 1979. However, building or programming a robotic hand that can successfully grasp different objects with a variety of shapes and sizes still remains a challenge in both the industry, and academia. This is an unmet need, especially in industries where robots are required to grasp numerous geometrically different objects in a dynamic environment. Several different approaches have been tried and tested for this problem, with varying degrees of success. However, the recent boom in Deep Learning and Artificial Intelligence in general has allowed for a new avenue to be pursued in academia. The idea is to train the robot to grasp objects by training a Neural Network, that will learn the optimum strategy – gripper angle, gripper height, angle of the arm, etc – to grasp an object of that particular shape and size.\\

A research paper recently proposed the idea of a self-supervised robot which labels and marks successful attempts of grasping in the process of training the CNN. However, the problem of limited datasets persists, as larger sets of training for a robot means better performance in the testing phase where an unidentified new object is presented to the robot for grasping. The probability of successful grasping increases with more training. The setup for this robot however is expensive, large and power hungry. We propose a much smaller hand-eye coordination based self-supervised robot which can train a CNN for several hundred hours on multiple objects in a lab setup. Then this CNN can be scaled onto the larger robots or integrated with their present CNN. This way we will not only save resources but also recreate a setup otherwise too expensive and power-intensive for direct training.
\pagebreak

\section{Chapter 3: Literature Review}
\subsection{Recent Research}
In a paper published in 2016 by Sergey Levine, Peter Pastor, Alex Krizhevsky and Deirdre Quillen under the name ‘Learning Hand-Eye Coordination for Robotic Grasping with Deep Learning and Large-Scale Data Collection’, a learning-based approach to hand-eye coordination for robotic grasping from monocular images is described\cite{ak1}. A large convolutional neural network (CNN) was trained to predict the probability of the motion of the gripper that would result in a successful grasp. It was accomplished using between 6 and 14 robotic manipulators with minor differences in camera placement and hardware where 800000 grasp attempts were employed to train the CNN. The method consisted of two parts: a grasp success predictor which uses deep CNN to result in a successful grasp, and a continuous servoing mechanism that uses the CNN to continuously update the robots motor commands providing the robot with fast feedback to perturbations and object motion. The main contribution of this work was a method for learning continuous visual servoing for robotic grasping, a novel CNN architecture for predicting the grasp success, and a large-scale data collection framework for robotic grasps.
\subsection{Research in Context}
The work here is similar to that by Pinto \& Gupta (2015)\cite{ag1} except that the latter used the open-loop approach. The open-loop approach does not employ continuous visual servoing but instead observes the scene prior to the grasp, extracts image patches, chooses the patch with the highest probability of a successful grasp, and then uses a known camera calibration to move the gripper to the location. More work includes Weisz \& Allen (2012) and (Rodriguez et al., 2012) which analyze the shape of the target object first and then plan a suitable grasp pose. Work of (Goldfeder et al., 2009) uses depth or stereo sensors to estimate the geometry of the scene and matching of previously scanned models to observations. Works that take on the data driven approach to predict grasp configurations include (Herzog et al 2014; Lenz et al 2015). There are also significant number of works that cannot be included in this report but the problem with what has been done so far is that the data points are not scalable for industrial scale use.\\
\section{Chapter 4: Problem Statement}
\subsection{Problem Elaboration}
Many approaches traditionally used human-labeled data to train the Neural Network in question. However, all of them faced problems during the training mechanism, as the data acted as a bottleneck in this approach. Labeling will always be biased due to semantics involved, and there can be multiple approaches to grasp a certain object. Human labeling is also a tedious and time consuming work. All of these reasons mean that ‘human generated data’ is not a scalable solution.\\
\subsection{Impact of AI on the environment}
There has been a lot of development in the world of deep learning and AI recently, with a lot of research being done on these new, emerging ideas. However, these new developments are not without their cost. The latest Deep Learning frameworks require huge amounts of computational resources to train. In a study conducted by OpenAI, since 2012, the amount of computational resources required for AI training have increased exponentially, with a doubling period of 3 to 4 months (in comparison, Moore's Law had a doubling period of about 2 years)\cite{oa1}. In another incredibly shocking study published in the MIT Technology Review, training one large AI model could produce as large of a carbon footprint as five cars in their entire lifetimes\cite{mt1}. All of these studies point to the fact that AI training has started taking a huge toll on the environment. It is becoming a huge problem, that needs to be addressed by the community at large.\\

The research community has been pushing for these advances in AI and Machine Learning for years, with little regard to what it costs. The cost of training AI models is more than just monetary. The energy consumption of a single model's training process is becoming hugely problematic. The toll this push in research without thinking about the consequences is taking on the environment, is a problem that researchers will have to think about deeply moving forward. Scalable and smaller, lower cost setups with low energy consumption and smaller carbon footprints hence become extremely important aspects of any AI research.\\

Robotic grasping is still an active research area, and a lot of strides have been made in recent years to setting up a self-supervised robotic grasper, that automates the collection of data. However, nearly all of these efforts are done on extremely expensive hardware that consumes huge amounts of energy in the process. If the training process is run directly on these types of hardware, the carbon footprint of such a research effort rises exponentially. None of the robotic grasping setups proposed so far are small, low cost, scalable, and have low energy consumption. In other words, none of the robotic setups proposed so far address the negative impacts of AI training on the environment.\\
\subsection{Problem Definition}
The problem this project aims to solve, is \textbf{the elimination of human involvement – data labeling and human bias – from the training process of robotic grasping}. This means that the main objective of this project will be to \textbf{create ‘Self Supervised Robots’, that can effectively solve the robotic grasping problem using minimal amount of human involvement}, while also keeping in mind the usual engineering problem of minimizing energy consumption and reducing equipment costs.
\subsection{Practical Applications}
While this is in essence a research project, a successful and scalable solution to the robotic grasping problem will have a huge impact both in academia and in industry. In academia, the solution would mean a push towards research in areas such as automated industrial processes control, system design for material processing, and human motor control etc\cite{tb3}. However, it is the industry that will perhaps be the biggest benefactor of the solution of this problem. Fast growing industries such as E-Commerce have a huge demand of robots that are able to grasp a variety of objects in order to fully automate the process. Robots are providing far greater flexibility and dramatically expediting the Return on Investment (ROI) of warehouse automation\cite{tb4}. Other industries that will greatly benefit from a solution to this problem include the food industry (automating food packaging and processing), toy industry (grasping and classifying plastic parts), storage and logistics applications (automated stocking and transportation of products), and automotive manufacturing (robotic assemblies). Other indigenous projects that will benefit from such a robot include automating systems such as litter-picking robots, and automated robots in the service industry to name a few\cite{tb5}.
\pagebreak
\section{Chapter 5: Project Overview}
\subsection{Overview of Research Project}
The overall goal of this project, as mentioned earlier, is to minimize and eliminate the need for human interaction with the robots and super-size the idea of ‘Self-supervised Robots’. Previous works have shown a promising future towards Robot-Object interaction with advent of DL and AI but none of the approaches is scalable and up-for-grabs for industries. Our proposed method involves using ROS (Robot Operating System) on numerous LEGO kits in a closed lab environment where the real time interaction of the robotic kits with the objects would train the CNN if a grasp was successful or not. The data points generated out of these hours of robot-object interaction would then be made available for large industrial-scale robots with extended accuracy and lower failure risks. The breaking down of learning process to numerous trainer robots curtails the input (energy consumption, equipment cost, time investment) and maximizes the output labeled data set.
\subsection{Research Project Goals}
Given the success of the project, the hardware components would include trainer robot kits\cite{tb1} that would be installed in industry environment and trained on concerned objects to create novel data points. This expanding cloud of AI labelled data set would then be used by the end-effector robots that actually perform the automation tasks. The packaged software would include ROS\cite{tb2} as it is becoming increasingly popular to program robots. In a nutshell, the proposed project promises a complete package for industrial automation, probably the first time where the concept of what started with robot grasping now shakes hands with the industry.
\subsection{Specifics of this Project}
While the overall research project involves both hardware and software components, due to the pandemic, these tasks had to be split. Hence, the goal of this particular Final Year Project was to develop the software end of this overall robotic grasping setup. This meant that the major tasks of this project included developing an object detection framework, integrating the CNN to obtain grasp predictions, generation of data-points to form a larger dataset, and training of the CNN on that dataset for better robotic grasping.
\section{Chapter 6: Project Development Methodology}
The overall research development was broken into two distinct parts - the hardware for robotic grasping, and the software for grasp angle prediction. This Final Year Project focuses on the software aspects of the robotic grasping problem.
\subsection{Software-Hardware Communication}
\begin{figure}[H]
\centering
\includegraphics[scale=0.6]{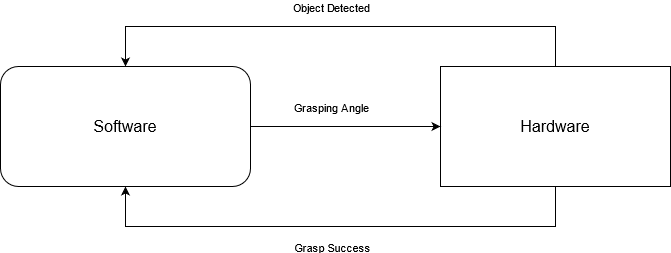}
\caption{Overall Project Overview}
\end{figure}
\bigskip
Figure 1 details the overall project. The hardware and software modules will have to communicate over several aspects. Specifically, whenever the software receives a signal from the robot that an object has been detected, the software has to produce a prediction of the optimal grasp angle for the object. This grasp angle should then be sent to the hardware, where the grasp should be executed. After the execution of the grasp, the feedback i.e whether the grasp was successful or not should be sent back to the software node for datatset generation.
\subsection{Details of Modules}
\begin{figure}[H]
\centering
\includegraphics[scale=0.6]{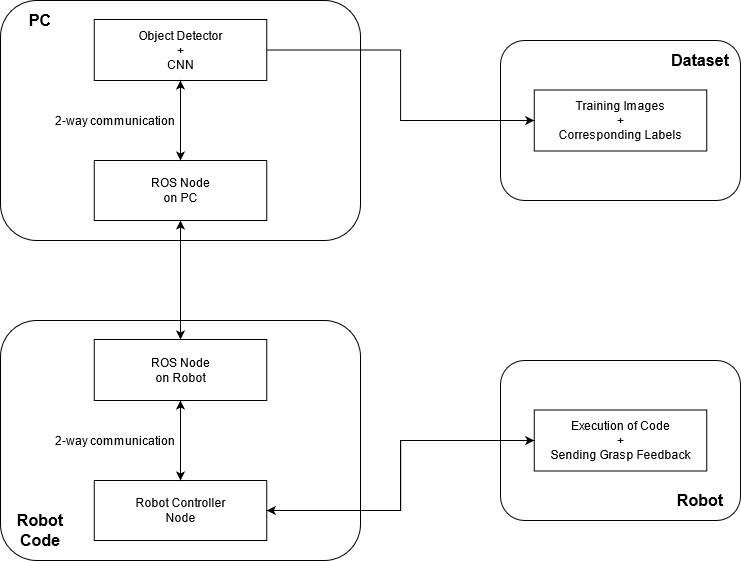}
\caption{Detailed Project Modules}
\end{figure}
\bigskip
Figure 2 outlines the details of this project on both the hardware and software side. For communication between the robot and the PC, we use ROS. The Robot Operating System or ROS offers a very convenient way of communication. In ROS, we have nodes and topics. ROS Nodes can publish information on ROS Topics, and can also receive information from them. This means that different ROS Topics can be created for the communication of various pieces of information, making information quick and easy to access.\\

On the PC side, an Object Detector and a Convolutional Neural Network (CNN) will be up and running. As soon as the signal of an object being detected is sent to the ROS Node on PC, it communicates this information to the `Object Detector + CNN' Node. The Node then takes a picture of the object, and passes this image through the CNN. The CNN outputs a prediction of the optimal grasp angle for the object. This grasp angle is sent to the ROS Node on PC, which in turn sends it to the Robot where the grasp is executed.\\

After the grasp has been executed, the success of the grasp is sent by the robot to the ROS Node on PC once again, where this information is communicated back to the `Object Detector + CNN' Node. The object's image along with its corresponding labels - predicted grasp angle and the grasp success - will form a complete data-point for our dataset. This information is saved on each grasp attempt, leading to the formation of a larger dataset that can be used to train not only this particular robot, but any other robot's Convolutional Neural Network as well.\\

The next section details the project design. As this project focuses on the software aspects of this overall research setup, we will mostly be focusing on the \textbf{PC} and the \textbf{Dataset} modules. We will treat the hardware as a generic block, considering only its inputs and outputs rather than its inner working. For the details on the working of the hardware, you can take a look at the project report \textbf{``Low Cost Robotic Grasper"} by Salman Nasir, Muhammad Obaid-ur-Rahman, and Abdul Mohaiman Hashmi.
\pagebreak
\section{Chapter 7: Detailed Design and Architecture}
The complete code of the project can be accessed at: \url{https://github.com/danyalsaqib/self-supervised-robotic-grasping}\cite{ds}
\subsection{The GUI}
The Graphical User Interface or GUI was made in Python using the library `Tkinter'. The following snippet of code details the design of the GUI:
\begin{lstlisting}[style = pycode, title = {\textbf{Code: \href{https://github.com/danyalsaqib/self-supervised-robotic-grasping/blob/main/PC Files/functs.py}{functs.py}}}]
		# Button for starting object detection
        self.lol1 = 0        
        self.controlVar = False
        self.btn = Button(self.root, text="Click to start Object Detection", 
                          command=self.changeText, height=3, width=30)
                          
        self.btn.config(bg = '#DCDCDC')
        self.btn.grid(row = 7, column = 5, pady = 2,
                      columnspan = 1)
        
        # Button for capturing image
        self.controlVar2 = False
        self.btn2 = Button(self.root, text="Capture Image", 
                          command=self.captureImage, height=3, width=30)
                          
        self.btn2.config(bg = '#DCDCDC')
        self.btn2.grid(row = 4, column = 5, pady = 2,
                      columnspan = 1)
        
        # Box that shows detected object
        self.labela1 = tk.Label(self.root, text = "Detected Object: ",
                                fg = 'white')
                                
        self.labela1.config(bg = '#565656')
        self.labela2 = tk.Label(self.root, textvariable=self.avar,
                                 height=2, width=15)
                                 
        self.labela2.config(bg = 'white')
        self.labela1.grid(row = 0, column = 4, sticky = W, pady = 10) 
        self.labela2.grid(row = 0, column = 6, sticky = W, pady = 10,
                          columnspan = 2) 
\end{lstlisting}
\bigskip
\indent As can be seen from the above code, we define two buttons for our GUI. One, when clicked, will turn on any attached camera and start the object detection process (which will be detailed later). The second button is used for capturing an image. When clicked, it will take a picture using the camera, and save it in the specified directory. A clever piece of coding means that the second button only appears when the `Click to start Object Detection' button has already been clicked. This means that the option to save the image only appears when we are actually running the object detector. While the first button will always be need to be clicked to start the object detection, as will be seen later, we will actually automate the process of saving the image for this setup.\\
\begin{figure}[H]
\centering
\framebox{\includegraphics[scale=0.6]{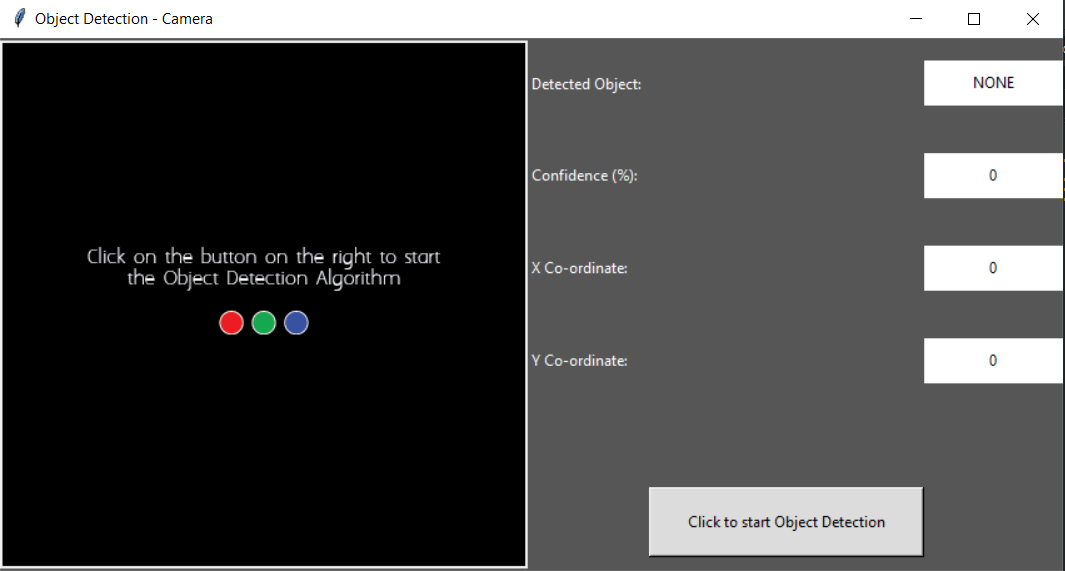}}
\caption{Object Detector GUI}
\end{figure}
\bigskip
The image of the final output of the GUI can be seen in figure 3. Along with a box that details the object being detected, boxes were also added that would detail the object's confidence percentage along with the exact x-coordinate and y-coordinate. While the hardware available to us was quite simple and did not factor in the object's x and y coordinate while grasping, it may be useful when extending the project to take these factors into account.
\pagebreak
\subsection{The Object Detector}
While there were quite a few choices when it came to the object detection architecture, we ultimately made the choice to go with the YOLO framework\cite{tb7}. At the time of writing, YOLOv4 is the latest version available, and hence it is chosen to be our framework of choice. We use the pretrained YOLOv4 architecture, that has been trained on the COCO (Common Objects in COntext) Datatset.
\begin{lstlisting}[style = pycode, title = {\textbf{Code: \href{https://github.com/danyalsaqib/self-supervised-robotic-grasping/blob/main/PC Files/OD_Script.py}{OD\_Script.py}}}]
classesFile = 'coco_names.txt'
classNames = []

with open(classesFile, 'rt') as f:
    classNames = f.read().rstrip('\n').split('\n')
    
# print(classNames)
# print(len(classNames))

modelConfiguration = 'yolov4.cfg'
modelWeights = 'yolov4.weights'

net = cv2.dnn.readNetFromDarknet(modelConfiguration, modelWeights)
net.setPreferableBackend(cv2.dnn.DNN_BACKEND_OPENCV)
net.setPreferableTarget(cv2.dnn.DNN_TARGET_CPU)
\end{lstlisting}
\bigskip
\indent The snippet of code shown here outlines the import of the YOLOv4 architecture for object detection. You can see that at first, a list of all possible objects from the COCO dataset is imported as a txt file. The YOLOv4 weights and configuration files are then imported, and a Deep Neural Network is then defined using the cv2 library.
\begin{figure}[H]
\centering
\framebox{\includegraphics[scale=0.6]{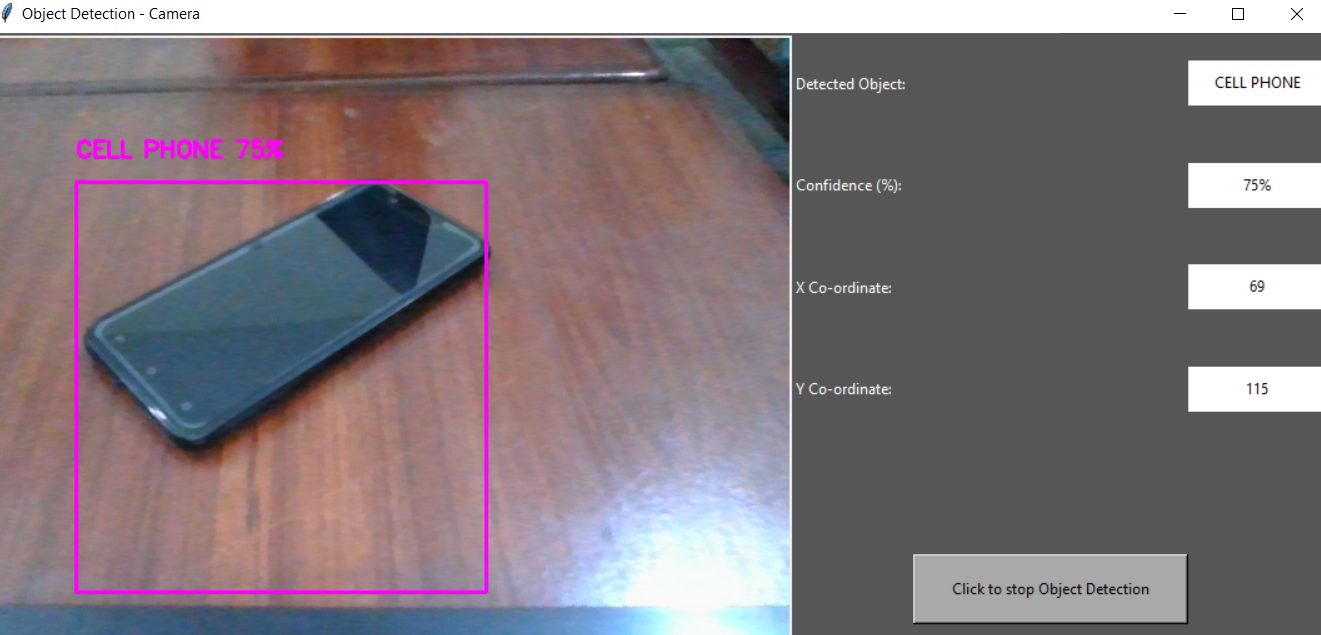}}
\caption{YOLOv4 Object Detector with GUI}
\end{figure}
\bigskip
\begin{figure}[H]
\centering
\framebox{\includegraphics[scale=0.6]{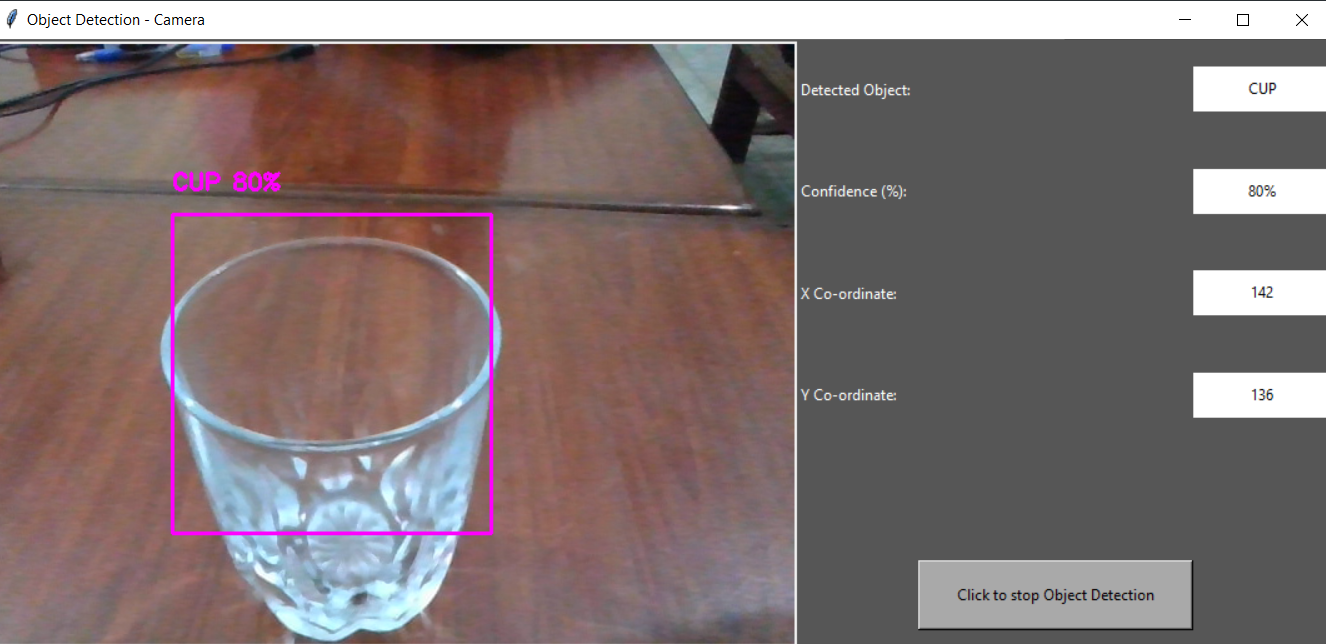}}
\caption{Object Detection on various objects}
\end{figure}
\pagebreak
\subsection{The Convolutional Neural Network}
The Convolutional Neural Network or the CNN is at the heart of the whole project. Neural Networks have long been a key component of computer vision tasks, and this one is no different. The CNN's function within this framework is to take images of objects taken by the camera as its input, and output a prediction of the optimal grasp angle ranging between 0 degrees and 180 degrees.
\subsubsection{The CNN Architecture}
While the problem seems quite straightforward at first, there is a crucial decision to be made when deciding on the CNN's architecture. There are 2 distinct possibilities when our desired output is an angle between 0 degrees and 180 degrees. The CNN's output could be a continous function by treating this as a regression problem. This way, the output would be defined as a single function. Alternatively, this could be treated as a classification problem, with 18 different classes, and each class corresponding to a grasp angle 10 degrees apart. These 2 opposing architectures will both work for our task at hand, but the training process will differ significantly for both. Thus, this is a critical decision to make. Here, we quote an extract from Lerrel Pinto and Abhinav Gupta's research paper on the subject:\\\\
\noindent \textbf{``One can train the grasping problem as a regression problem: that is, given an input image predict $\bm{(x, y, \theta)}$. However, this formulation is problematic since: (a) there are multiple grasp locations for each object; (b) CNNs are significantly better at classification than the regressing to a
structured output space" ... ``given an image patch we estimate
an 18-dimensional likelihood vector where each dimension
represents the likelihood of whether the center of the patch is graspable at $\bm{0^\circ}$, $\bm{10^\circ}$, ... $\bm{170^\circ}$. Therefore, our problem can be thought of an 18-way binary classification problem."}\cite{ag1}\\

Hence, we go for the classification approach rather than a regression approach. So our CNN will take an image as an input, and will output likelihoods for all 18 classes for a grasp to be successful. The class with the highest likelihood will then be selected as the chosen optimal grasp angle.
\pagebreak
\subsubsection{The CNN Model}
After the architecture has been decided, we now have to actually define the CNN model. Training the CNN from scratch is a lengthy and tedious task, that is extremely time consuming and requires a lot of computational resources. Hence it makes sense to start off from a pre-trained model. While there are several pre-trained models available, we choose the ResNet model. The reasons for choosing this particular model are quite clear. ResNet is the latest in a list of the state-of-the-art Neural Network architectures. Not only this, but the `skip connections' in a ResNet model ensure that we do not encounter the vanishing gradient problem along with a faster training time. All of these points affirm our choice to use a ResNet architecture.
\begin{lstlisting}[style = pycode, title = {\textbf{Code: \href{https://github.com/danyalsaqib/self-supervised-robotic-grasping/blob/main/PC Files/functs.py}{functs.py}}}]
model_ft = models.resnet18()
num_ftrs = model_ft.fc.in_features
model_ft.fc = nn.Linear(num_ftrs, 18)
model_ft.load_state_dict(torch.load('fyp_cnn.pt'))
\end{lstlisting}
\bigskip
\indent While Tensorflow and Keras were also readily available choices for our CNN-related tasks, we choose PyTorch as the library of our choice. As can be seen from this snippet of code, the variant of ResNet we are using is ResNet18. The ResNet18 model is loaded, and the CNN architecture is then modified so that the last layer has 18 different classes, each corresponding to a grasp angle. Lastly, the model weights are loaded from the file titled `fyp\_cnn.pt'. This way, whenever we train our model on our collected data, we can easily load the newly trained weights here.
\pagebreak
\section{Chapter 8: Implementation of Modules}
\subsection{Signal - Object Detection}
As said earlier, the hardware of the project will be treated as a black box. As such, we will talk about our implementation purely in terms of the software. Firstly, let's look at the code that is run when an object is detected.
\begin{lstlisting}[style = pycode, title = {\textbf{Code: \href{https://github.com/danyalsaqib/self-supervised-robotic-grasping/blob/main/PC Files/OD_Script.py}{OD\_Script.py}}}]
    success, img = cap.read()
    img = cv2.flip(img, 1)
    obj_stat = open("objectstatus.txt", "r")
    os_var = int(obj_stat.read())
    if os_var == 1:
    		wind.controlVar2 = True	
\end{lstlisting}
\bigskip
\indent The code shows that an image is being updated into the `img' variable. Then, the file `objectstatus.txt' file is read. If \textbf{`1'} is written into the file, then the hardware has signaled that an object has been detected near the grasper. Hence, if there is 1 written into the file, a control variable `wind.controlVar2' has the \textbf{`True'} boolean value written to it.
\subsection{Image Saving and CNN Prediction}
After the object has been detected, the next steps are to save an image of the object, and obtain a CNN prediction of the optimal grasp angle.
\begin{lstlisting}[style = pycode, title = {\textbf{Code: \href{https://github.com/danyalsaqib/self-supervised-robotic-grasping/blob/main/PC Files/OD_Script.py}{OD\_Script.py}}}]
if wind.controlVar2:
        global gc
        # Change Directory to where you want your images saved
        os.chdir('/home/dani/catkin_ws/src/tutorials/CNN Training/Training Images')
        print("Image Write Success: ", cv2.imwrite(os.path.join('train_img_'+str(gc)+'.png'), img))
        os.chdir('/home/dani/catkin_ws/src/tutorials')
        cv2image = cv2.cvtColor(img, cv2.COLOR_BGR2RGBA)
        im = PIL.Image.fromarray(cv2image) 
        im1 = im.convert('RGB')		# Image converted to RGB
        img_nn_mod = data_transforms(im1)	# Appropriate transforms applied to image
        img_nn_mod = img_nn_mod.unsqueeze(0)  # if torch tensor
        outputs = model_ft(img_nn_mod)	# Image being passed through CNN
        _, preds = torch.max(outputs, 1)	# Obtained Predicted Class
        angle = int(preds) * 10
        angle = angle - 90
        angle = angle / 90
        angle = angle * 0.25
        print("Prediction Class:", int(preds) * 10)
        print("Mapped Output: ", angle)
        f = open("cnnoutput.txt", "w")
        f.write(str(0.5) + "\n")
        f.write(str(angle))
        f = open("gc_file.txt", "w")
        f.write(str(gc))
        obj_stat = open("objectstatus.txt", "r")
        os_var = int(obj_stat.read())
        while os_var == 1:
        		obj_stat = open("objectstatus.txt", "r")
        		os_var = int(obj_stat.read())
        		obj_stat.close
        		time.sleep(3)
\end{lstlisting}
\bigskip
\indent The code above details the steps that take place when `wind.controlVar2' is true i.e an object has been detected. First, an image of the object is saved into a pre-defined directory named `/home/dani/catkin\_ws/src/tutorials/CNN Training/Training Images'. After the image has been saved, the image has the appropriate transforms applied to it. Firstly, the image is converted from a cv2 array to an image format. Then the image is converted into RGB, and the appropriate torch transforms are applied.\\

Now, the image is ready to pass through the CNN. The image is passed through the CNN (model\_ft), and the output of the CNN, which is a tensor of size 18, is saved into the variable outputs. From this tensor, the tensor index with the highest value (corresponding to the highest likelihood of successful grasping) is extracted, and saved into the `preds' variable. Simply multiplying this value by 10 gives us the exact angle required by the robotic grasper. This angle value is first mapped according to the robot's motors, and it is then written to the file `cnnoutput.txt'.\\

Now, the program waits for the `objectstatus.txt' to have 0 written into it. This is because when 0 is written to `objectstaus.txt', it is a signal to the program that the grasp has been executed and the feedback of the grasp has been successfully sent back to the PC.
\subsection{Grasp Feedback Management}
After the grasp has been successfully executed, we have to use the feedback sent by the robot to the PC to form a complete data-point.
\begin{lstlisting}[style = pycode, title = {\textbf{Code: \href{https://github.com/danyalsaqib/self-supervised-robotic-grasping/blob/main/PC Files/OD_Script.py}{OD\_Script.py}}}]
        f = open("graspfeedback.txt", "r")
        rnm_var = int(f.read())
        os.chdir('/home/dani/catkin_ws/src/tutorials/CNN Training/Training Images')
        os.rename(os.path.join('train_img_' + str(gc) + '.png'), os.path.join(str(rnm_var) + '_' + str(int(preds))+ '_train_img_' + str(gc) + '.png'))
        os.chdir('/home/dani/catkin_ws/src/tutorials')

        gc = gc + 1
        print("controlVar2 exiting")
        wind.controlVar2 = False
\end{lstlisting}
\bigskip
\indent The feedback of the attempted grasp is written to the `graspfeedback.txt' file, with `1' signaling that the grasp was successful, and `0' signaling an unsuccessful grasp. This information now has to be stored in a way that makes it clear that this grasp result is for the previously saved image. Instead of making a separate spreadsheet or xlsx file, we directly encode this information into the previously saved image name. As seen in the code, the grasp feedback is read into a variable, and the previously renamed file has the grasp success, and the angle of attempted grasp appended to its name as a prefix.
\pagebreak
\section{Chapter 9: Complete Integration and Testing}
While this particular project focuses on software, to complete the setup, we also have to setup the hardware. Hence, this section will give an overview of what to do to completely setup the whole project.
\subsection{Hardware Setup}
The LEGO EV3 kits are set up in such a way that we work with 3 motors, each connected to a port of the EV3 brick. One port of the EV3 is also connected with an ultrasonic sensor. The following mapping outlines how each of these components are connected.
\begin{itemize}
\item EV3 Brick Port A $\rightarrow$ Motor for moving claw up and down
\item EV3 Brick Port B $\rightarrow$ Motor for rotating claw
\item EV3 Brick Port A $\rightarrow$ Motor for opening and closing jaws
\item EV3 Brick Port 1 $\rightarrow$ Ultrasonic sensor for object detection
\end{itemize}
\indent Apart from the robot itself, a webcam is needed to act as the robot's eyes. While a laptop's built in webcam may theoretically work, we recommend working with a USB webcam, that can be set up appropriately so as to be able to clearly view the object.
\begin{figure}[H]
\centering
\framebox{\includegraphics[scale=0.45]{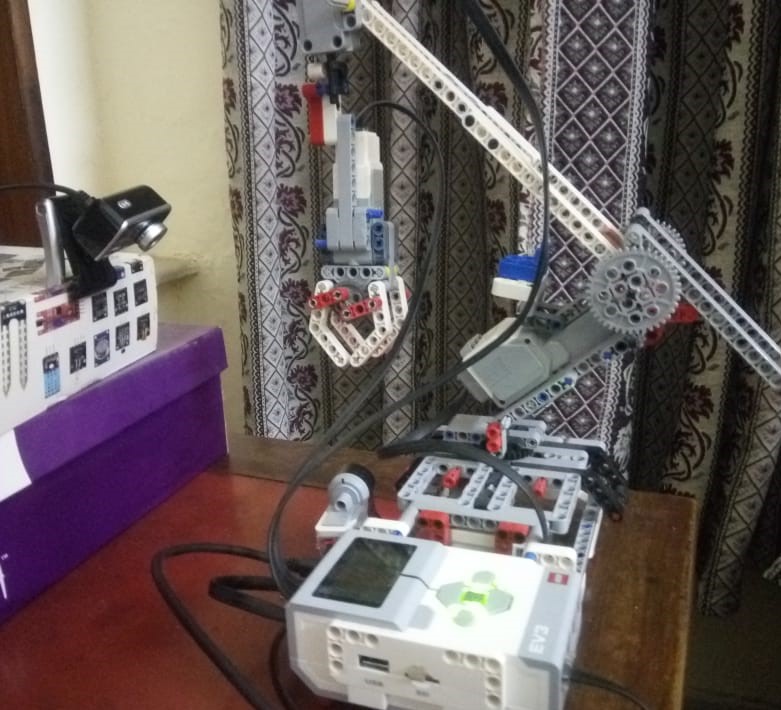}}
\caption{The Hardware Setup}
\end{figure}
\subsection{Software Setup and Installation}
The project involves 2 nodes - the PC and the LEGO EV3 Robot. Both of them require ROS to be installed on them. Our particular setup had Debian Linux installed on the EV3 LEGO Brick, and ROS Noetic on our Ubuntu Linux Machine. We will assume in this walkthrough that ROS is installed on both the EV3 Lego kit and on your PC. You can follow the links below for thorough guidance:
\begin{itemize}
\item Installing ROS on a LEGO EV3 Brick: \url{http://wiki.ros.org/Robots/EV3}
\item Ubuntu install of ROS Noetic: \url{http://wiki.ros.org/noetic/Installation/Ubuntu}
\item Installing Catkin: \url{http://wiki.ros.org/catkin}
\end{itemize}
\indent After ROS has been installed on both the machines, the next step is to make a new catkin workspace. On your PC, open up the terminal and run the following commands line by line:
\begin{lstlisting}[style = bash_code]
$ mkdir -p ~/catkin_ws/src
$ cd ~/catkin_ws/
$ catkin_make
\end{lstlisting}
\bigskip
\indent After running these commands, you should have a new catkin workspace that has various subdirectories. To set the files up on PC, copy all of the files from the 'PC Files' folder of this repository, and copy them into the directory 'catkin\_ws/src/tutorials'. Create this directory if it hasn't already been created using the following commands:
\begin{lstlisting}[style = bash_code]
$ cd catkin_ws/src
$ mkdir tutorials
\end{lstlisting}
\bigskip
\indent After all of the files have been copied to this directory, run the following lines on the terminal again:
\begin{lstlisting}[style = bash_code]
$ cd ~catkin_ws
$ catkin_make
\end{lstlisting}
\bigskip
\indent Hopefully, we should be all set up now on the PC. We will now move on to setting stuff up on the robot.\\

\indent Firstly, to access the robot via any one of your PC terminals, connect the LEGO EV3 brick to your PC via the USB cable, and run the following command from your Ubuntu machine:
\begin{lstlisting}[style = bash_code]
$ ssh robot@ev3dev.local
\end{lstlisting}
\bigskip
\indent You will now be asked to enter a password. This password is maker by default, but you may have changed it for your particular setup. Enter the password, and you will be able to access the robot's files. Here, again run the following commands:
\begin{lstlisting}[style = bash_code]
$ mkdir -p ~/catkin_ws/src
$ cd ~/catkin_ws/
$ catkin_make
\end{lstlisting}
\bigskip
\indent Now, you will have to either copy the 'Robot Files' folder of this repository, and copy them into the directory 'catkin\_ws/src/tutorials/scripts'. If you cannot directly copy these files into the robot, simple create these files manually in the directory using the `nano command', for example `nano reset.py', and copy the text from this github repository's files into these manually created files. Create this directory if it hasn't already been created using the following commands:
\begin{lstlisting}[style = bash_code]
$ cd catkin_ws/src
$ mkdir tutorials
$ cd tutorials
$ mkdir scripts
\end{lstlisting}
\bigskip
\indent After all of the files are successfully copied onto the robot, run the following command:
\begin{lstlisting}[style = bash_code]
$ cd ~catkin_ws
$ catkin_make
\end{lstlisting}
\pagebreak
\subsection{Running the Setup}
To run the complete setup, 5 separate terminals are needed. We will be running 4 different python files on 4 different terminals, and the 5th terminal will be running the roscore.
\subsubsection{Terminal 1}
On the first terminal run the following command:
\begin{lstlisting}[style = bash_code]
$ roscore
\end{lstlisting}
\indent On this terminal, the roscore will start running. The roscore is what enables communication between the robot and the PC. Leave the roscore running on this terminal.
\subsubsection{Terminal 2}
On this terminal, run the following commands:
\begin{lstlisting}[style = bash_code]
$ cd ~catkin_ws/src/tutorials
$ python3 OD_Script.py
\end{lstlisting}
\indent A GUI window will popup. Click the 'Click to start Object Detection' button on the GUI, which will open your webcam and start running the \textbf{YOLOv4 object detector} on your webcam. This will also initiate the CNN prediction for objects that will be grasped. Again, leave this terminal running up.
\subsubsection{Terminal 3}
On this terminal, run the following commands:
\begin{lstlisting}[style = bash_code]
$ cd ~catkin_ws/src/tutorials
$ rosrun tutorials nodeA.py
\end{lstlisting}
\indent This will start up the ROS node on your PC for communication with your EV3 robot setup. Leave this terminal running as well.
\subsubsection{Terminal 4}
On this terminal, you will have to enter the robot using the following command:
\begin{lstlisting}[style = bash_code]
$ ssh robot@ev3dev.local
\end{lstlisting}
\indent Enter the password to access the robot. After the robot has been successfully accessed, run the following commands:
\begin{lstlisting}[style = bash_code]
$ cd ~catkin_ws/src/tutorials/scripts
$ python3 sample.py
\end{lstlisting}
\indent This will run the 'Robot Control Center' on this terminal. What this means is that all of the robot's instructions and movements will be regulated via this terminal or node. Leave this terminal running again.
\subsubsection{Terminal 5}
On this terminal, you will once again have to enter the robot using the following command:
\begin{lstlisting}[style = bash_code]
$ ssh robot@ev3dev.local
\end{lstlisting}
\indent Enter the password to access the robot. After the robot has been successfully accessed, run the following commands:
\begin{lstlisting}[style = bash_code]
$ cd ~catkin_ws/src/tutorials/scripts
$ rosrun tutorials test.py
\end{lstlisting}
This will run the ROS Node on the LEGO EV3 Robot to communicate with the PC.\\

After all 5 nodes are running, the setup will look something like this:
\begin{figure}[H]
\centering
\framebox{\includegraphics[scale=0.33]{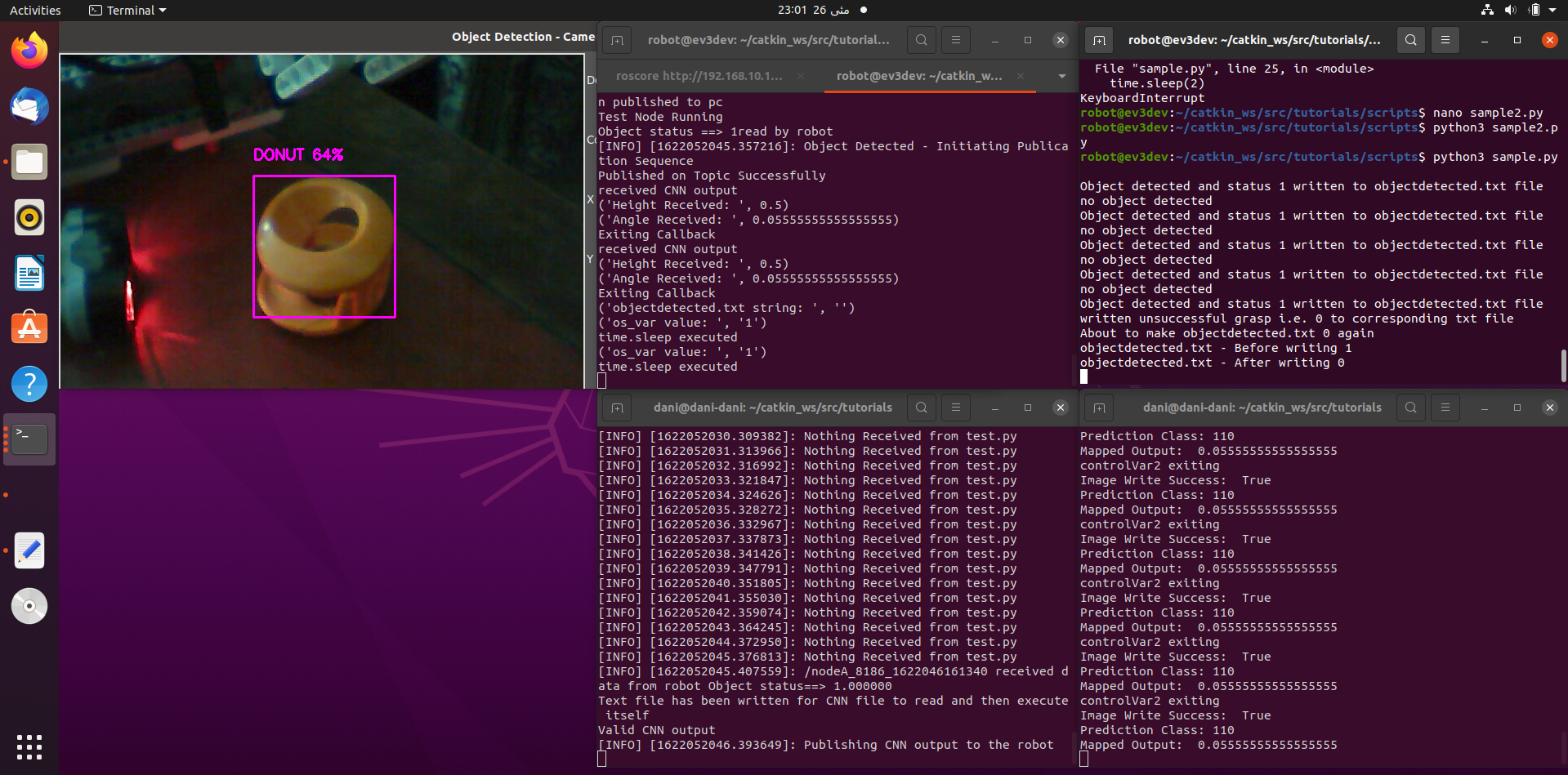}}
\caption{The Complete Software Setup}
\end{figure}
\section{Chapter 10: Results and Discussion}
\subsection{Dataset Multiplication}
After the setup has been completed, we are ready to generate a complete dataset for robotic grasping. We simply placed various objects in front of the robot, and because the program runs in a continuous loop, we can place objects and the robot will continue to generate new data-points. We just place new objects in front of the robot from time to time, and our dataset is successfully generated.\\

A demo video of the project in action can be seen at: \url{https://drive.google.com/file/d/16KXGa0TL62Kxv2mFmZ8jZYoCgTrSrEAR/view?usp=sharing}
\begin{figure}[H]
\centering
\framebox{\includegraphics[scale=0.4]{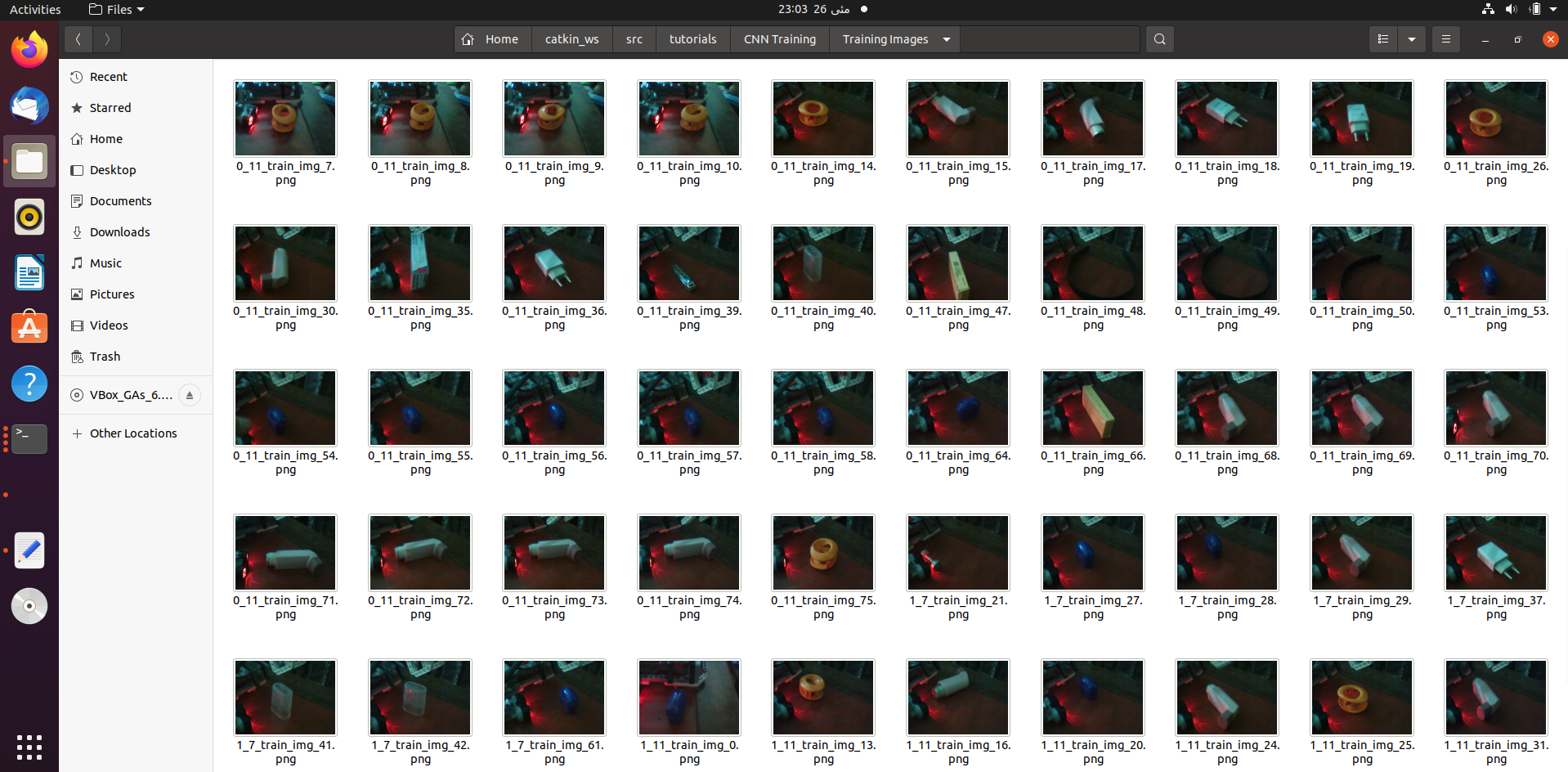}}
\caption{Generated Dataset}
\end{figure}
The figure shows the generated dataset from a session of robotic grasping. As can be seen from the image, a lot of data-points can be generated extremely quickly. This is is the advantage of such a self-supervised approach over human labeled datasets. Not only is novel dataset rapidly produced, but the dataset is also unbiased by semantics. In the modern era when data is becoming a bottleneck for AI and deep learning model's performance and quality data is hard to come by, this self-supervised grasping setup provides an elegant solution. Moreover, this particular project is performed using an extremely low cost setup, adding to its value.
\pagebreak
\subsection{Grasp Results}
\subsubsection{Images}
About 10 different objects were placed in front of the robot to grasp during the various sessions. The following section details some of the attempted grasps by the robot on various objects.
\begin{figure}[H]
\centering
\framebox{\includegraphics[scale=0.45]{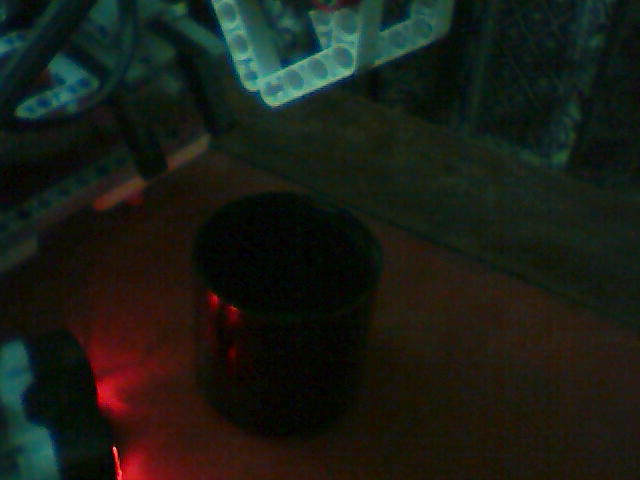}}
\caption{Object Grasp Angle: $70^{\circ}$, Grasp unsuccessful}
\end{figure}\begin{figure}[H]
\centering
\framebox{\includegraphics[scale=0.45]{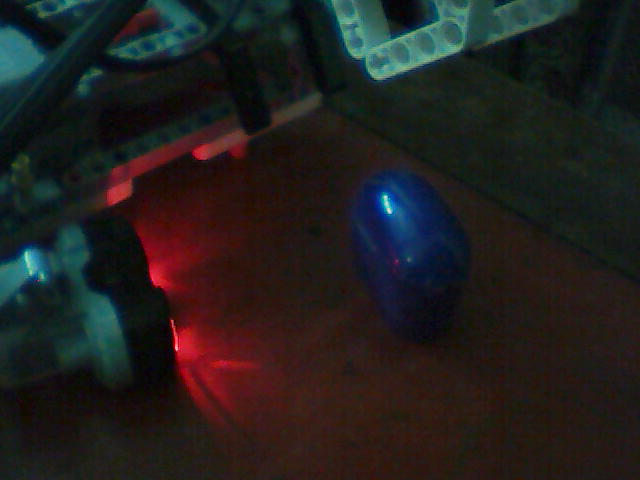}}
\caption{Object Grasp Angle: $110^{\circ}$, Grasp successful}
\end{figure}\begin{figure}[H]
\centering
\framebox{\includegraphics[scale=0.55]{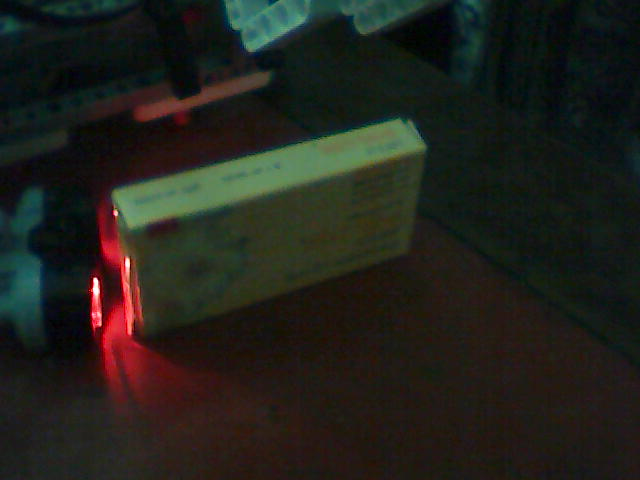}}
\caption{Object Grasp Angle: $30^{\circ}$, Grasp successful}
\end{figure}\begin{figure}[H]
\centering
\framebox{\includegraphics[scale=0.55]{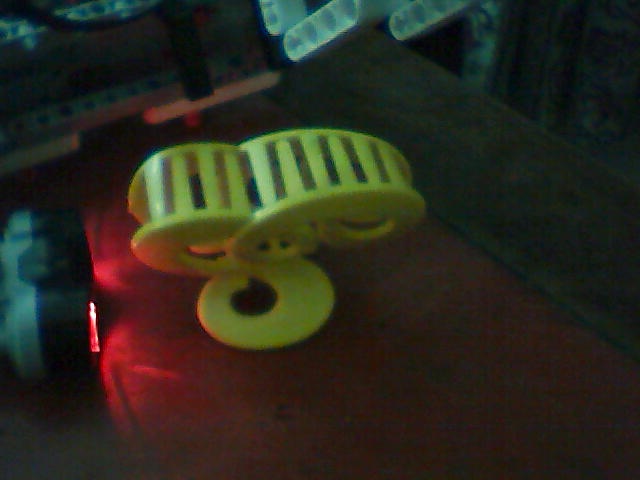}}
\caption{Object Grasp Angle: $110^{\circ}$, Grasp successful}
\end{figure}
\pagebreak
\subsubsection{Training on Dataset}
As a concept, a CNN was actually trained on this very generated dataset. While we did not have the resources to train CNN on our own machines, Google Colab presents us with a convenient solution. The 10 GB of free GPU storage is enough to train the ResNet based CNN used in our project.
\begin{figure}[H]
\centering
\framebox{\includegraphics[scale=1]{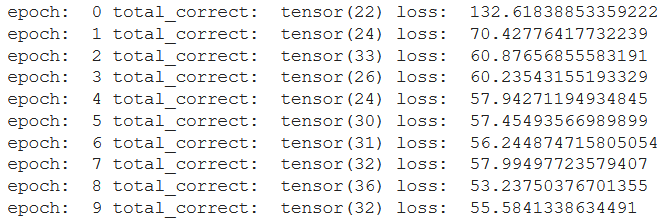}}
\caption{Training Results}
\end{figure}
As can be seen from the above figure, the training results show a constantly better performance with each epoch of the CNN training loop. Thus, we end up with a CNN that is better equipped to predict optimal grasp angles for a particular object. We can save this updated CNN from our Colab Notebook to our Google Drive using the following line of code:\\
\begin{lstlisting}[style = pycode]
torch.save(network.state_dict(), '/content/drive/My Drive/Final Defense/ fyp_cnn_updated.pt')
\end{lstlisting}
\bigskip
\indent This line of code saves the CNN to our Google Drive, where it can be downloaded and the older CNN in the setup can be replaced by this updated CNN by simply replacing the old `fyp\_cnn.pt' file by the newer `fyp\_cnn\_updated.pt' file. This completes our cycle, where this new CNN can now be used to generate newer data-points, and those newer data-points can in-turn be used to generate a better performing CNN.
\pagebreak
\subsubsection{Comparison of Grasp results}
The following tables outline the grasp results of the robot before and after training on the CNN.
\renewcommand{\arraystretch}{1.5}
\begin{table}[H]
\centering
\begin{tabular}{| p{0.3\textwidth} | p{0.3\textwidth} |}
 \hline
 \multicolumn{2}{| c |}{\textbf{Overall Statistics}}\\
 \hline
 \textbf{Successful Grasps} & 47\\ 
 \hline 
 \textbf{Unsuccessful Grasps} & 76\\ 
 \hline 
 \textbf{Total Grasps} & 123\\ 
 \hline	
\end{tabular}
\caption{Overall Grasp Results}
\end{table}
\begin{table}[H]
\centering
\begin{tabular}{| p{0.3\textwidth} | p{0.3\textwidth} |}
 \hline
 \multicolumn{2}{| c |}{\textbf{Session 1: Before CNN Training}}\\
 \hline
 \textbf{Successful Grasps} & 22\\ 
 \hline 
 \textbf{Unsuccessful Grasps} & 50\\ 
 \hline 
 \textbf{Total Grasps} & 72\\ 
 \hline
\end{tabular}
\caption{Grasp results before training}
\end{table}
\begin{table}[H]
\centering
\begin{tabular}{| p{0.3\textwidth} | p{0.3\textwidth} |}
\hline 
 \multicolumn{2}{| c |}{\textbf{Session 2: After CNN Training}}\\
 \hline
 \textbf{Successful Grasps} & 25\\ 
 \hline 
 \textbf{Unsuccessful Grasps} & 26\\ 
 \hline 
 \textbf{Total Grasps} & 51\\ 
 \hline
\end{tabular}
\caption{Grasp results after training}
\end{table}
\bigskip
\indent As can be seen from the above tables, the CNN training produces a significant boost in accuracy. The untrained CNN produces an accuracy of $30.55\%$, whereas the CNN after training produces an accuracy of $49.02\%$, a large improvement. This is the strength of this approach - each training cycle will improve the CNN's ability to predict optimal grasp angles.
\pagebreak
\section{Chapter 11: Conclusion and Future Work}
This project successfully implements a setup involving a self-supervised robot for novel dataset generation. Even though a few previous researches have already produced similar setups, none of them have been implemented at such a low cost. We used LEGO EV3 kits to make the hardware robotic setup. The CNN ran on a laptop with medium specifications, and the entire training process was done online using Google Colab's free cloud GPU storage. This means that this low cost solution is easily reproducible. However, the low cost solution comes with its limitations. The main constraints we faced were hardware-centric. The LEGO EV3 kit's robotic grasper is not very sophisticated, and is limited in what it can do. In addition, the number of motors had to be limited due to design constraints, limiting the degrees of freedom the robotic grasper could achieve. Hence, the cheaper setup does have its limitations.\\

What the low cost solution loses in accuracy and precision however, it gains in scalability. This completely self-supervised robotic setup is extremely scalable, and can be extended to a whole host of trainer robots. While due to the pandemic we were limited to running this robotic setup on a single trainer robot, future work could make use of our hardware and software developments to run this setup on several trainer robots simultaneously, boosting the amount of novel data being produced by these robots per unit time. The dataset generated could help in creating better Convolutional Neural Network architectures, that predict optimal grasping angles with greater accuracy. Moreover, this trained CNN could then be directly deployed in a larger robotic setup. The larger robot could not only give better results with the same CNN, but we would also save up significant costs. Instead of training directly on larger robots, training and dataset generation could be delegated to these smaller trainer robots, saving both money and valuable energy resources.\\

The successful completion of this project is an important achievement, and robotic grasping is still an extremely important area of research. Given the future prospects of this project and the venues for research, this achievement will hopefully lead to many more fruitful accomplishments in this domain.
\pagebreak

\end{document}